# Can LLMs Infer Personality from Real-World Conversations?

Jianfeng Zhu[1], Ruoming Jin [1], and Karin G. Coifman[2]
1 Department of Computer Science, Kent State University, Kent, OH 44224
2 Department of Psychological Sciences, Kent State University, Kent, OH 44224
*Jianfeng Zhu.
**Email:**  jzhu10@kent.edu

**Keywords:** Large Language Models, Personality Assessment, Big Five Inventory, Zero-shot Prompting, Chain-of-thought prompting, Psychometrics

## Abstract

Large Language Models (LLMs), such as OpenAI's GPT-4 and Meta's LLaMA, offer a promising path toward scalable personality assessment from open-ended language. However, inferring personality traits remains an inherently subtle task, and prior work has often relied on synthetic dataset or social media texts annotated via crowdsourcing, contexts that lack ecological and psychometric validity. To address this gap, we introduce a novel real-world conversational benchmark for evaluating LLM-based personality inference: a dataset of 555 semi-structured interviews eliciting autobiographical reflections on recent emotional and social experiences, paired with validated BFI-10 self-report scores collected in the same session. We evaluated three state-of-the-art LLMs (GPT-4.1 Mini, Meta-LLaMA, and DeepSeek) using zero-shot prompting for BFI-10 item-level prediction, and both zero-shot and chain-of-thought prompting for Big Five trait-level inference. While all models demonstrated high test–retest reliability across repeated runs, construct validity was limited: correlations with ground-truth scores were weak (maximum Pearson's r = 0.27), interrater agreement was low (Cohen's κ < 0.10), and predictions showed a decisiveness bias toward moderate-to-high trait levels. Chain-of-thought prompting and longer input contexts modestly improved distributional alignment but did not enhance trait-level accuracy. These findings underscore current limitations in LLMs' ability to infer validated personality constructs from naturalistic language, emphasizing the need for cautious, evidence-based development in psychological and clinical applications.

## Significance Statement

Personality traits shape cognition, behavior, well-being, making their accurate assessment essential for psychology, healthcare, and the development of responsible AI systems. Large Language Models (LLMs) offer scalable tools for infering personality from language, but existing evaluations have relied largely on synthetic or crowd-annotated data lacking ecological validity. We introduced the first benchmark grounded in real-world interview paired with validated self-report measures to evaluate LLM-based personality predictions. Although current models exhibit high internal consistency, they align poorly with gold-standard psychological assessments. These findings raise important concerns about their readiness for use in clinical, research, and applied settings. This work provides foundational insights into the challenges of current approaches and informs of more responsible, evidence-based AI for human-centered domains.

## Introduction

Personality, refers to enduring individual differences in characteristic patterns of thinking, feeling and behaving [1]. These differences encompass emotional tendencies, motivations, and values, that influence health, well-being, and interpersonal functioning. Accurate personality assessment is essential across domains such as healthcare settings [2,3], criminal justice [4,5], affective computing [6,7], and personalized AI systems [8–10]. Among theoretical frameworks, the Five Factor Model [11] is one of the most widely accepted and empirically supported approaches. It defines personality along five broad dimensions: extraversion, agreeableness, conscientiousness, neuroticism, and openness to experience [12]. These traits have been shown to predict a range of life outcomes, including interpersonal behavior, emotional experiences, work performance, and vulnerability to mental health conditions such as depression and anxiety [13,14].

Early computational methods for personality inference focused on digital footprints, notably Facebook likes, to predict personality traits [15,16]. Although these approaches showed promise, they raised significant ethical and privacy concerns, exemplified by the Cambridge Analytica scandal. More recent work has shifted towards publicly available textual data, such as tweets and social media posts, aiming to infer personality through natural language processing and machine learning techniques [17]. However, these methods typically rely on synthetic or crowd-labeled datasets with limited ecological or psychometric validity, reducing their generalizability to real-world applications [18].Traditional psychological methods for assessing personality primarily depend on self-report questionnaires, which, although validated, ca be time-consuming and resource-intensive [19,20]. This limits their scalability in large-scale or time-sensitive contexts.

The emergence of large language models (LLMs), including OpenAI's GPT-4 and Meta's LLaMA, has opened new possibilities for personality assessment. These models provides a wide range abilities, including serving as conversation agents, generating essays and stories, writing code, and diagnosing illness [21–25]. Accordingly, researchers have begun exploring whether LLMs can be used for psychological and personality assessment [26–28]. However, a critical question remains about: ***"Can LLMs Infer Personality from Real-World Conversations? "***.

To address this question, we evaluate three cutting-edge LLMs (GPT-4.1 Mini, Meta-LLaMA, and DeepSeek) on their ability to infer Big Five personality traits from naturalistic language. We introduce a novel benchmark: a dataset of 555 semi-structured interviews that elicited autobiographical reflections on recent emotional and social experiences, collected concurrently with validated Big Five Inventory (BFI-10) self-report scores [29,30]. Our methodological framework, including prompting strategies, evaluation metrics, and reliability analyses, is outlined in Figure 1.

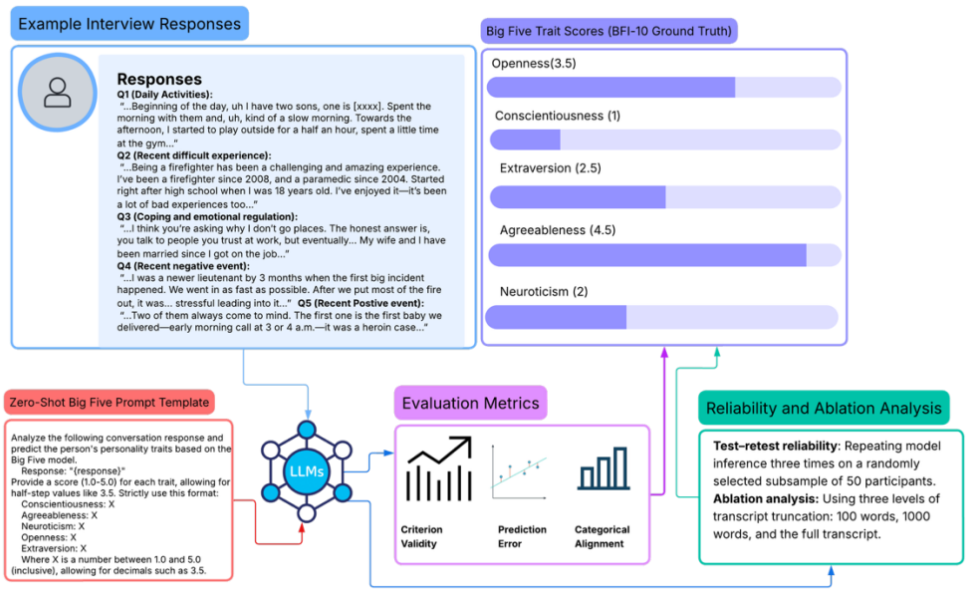

**Figure 1** Overview of the study design and methodology for evaluating large language model (LLM)-based personality inference.

We evaluate the models' performance using two prompting strategies, zero-shot prompting and chain-of-thought prompting. These approaches allow us to examine how different prompting formats influence predictive accuracy and psychometric alignment when using a 1–5 Likert scale [31]. Evaluation metrics include Pearson correlation coefficients, mean absolute error (MAE), root mean squared error (RMSE), and distributional alignment of discretized trait score. We further assessed prediction reliability across repeated runs and conduct an ablation study varying input length (100 words, 1000 words, and the full transcript) to evaluate the effect of context size.

This study investigates the following research questions:

**RQ1.** *How accurately can large language models (LLMs) predict validated BFI-10 personality scores from naturalistic, semi-structured interview transcripts?*

**RQ2.** *Does chain-of-thought prompting improve the accuracy and psychometric alignment of Big Five trait prediction compared to zero-shot prompting?*

**RQ3.** *How well do LLM-predicted trait distributions align with ground-truth binned distributions across the BFI-10 items and Big Five dimensions?*

**RQ4.** *How does input text length influence predictive accuracy, categorical reliability, and distributional alignment in Big Five trait inference?*

By addressing these questions, we aim to advance understanding of LLM capabilities for modeling human psychological traits and inform the design of future AI system that aspire toward greater personalization, emotional intelligence, and ecological validity in applied psychological settings and beyond.

## Results
### LLM Prediction of BFI-10 Trait Scores
Figure 2 and Table2 present a detailed comparison of model performance across the 10 BFI-10 items, evaluated using Pearson correlation, mean absolute error (MAE), and root mean square error (RMSE). Correlation values were generally low across all models and traits. However, relatively stronger correlation was observed for BFI-2 ("is generally trusting"), BFI-3("tends to be lazy"), BFI-7("tends to find fault with others"), and BFI-10 ("has an active imagination"). In contrast, BFI-5 ("has few artistic interests") and BFI-6 ("is outgoing, sociable") yielded consistently negative correlations across all models. Among the models, GPT-4.1 Mini showed the strongest alignment with self-reported scores, though correlations remained weak; its highest correlation was for BFI_3 (r = 0.27). The lowest correlation was observed for MetaLLaMA on BFI-5 (r=-0.18).

MetaLLaMA also produced the highest MAE and RMSE values, with mean errors approaching 2.0 across items. It performed worst on BFI_3 (MAE = 2.1; RMSE = 2.3) and BFI-7 (MAE = 2.0; RMSE = 2.3), indicating poor reliability even where weak correlations exsited. DeepSeek, despite showing the lowest average Pearson correlations (often near zero or negative), achieved the lowest MAE and RMSE score on BFI-2 and BFI-7-9 (MAE =1.2, RMSE ~=1.5).

Figure 3 visualizes predicted response distributions (1–5 scale) for each BFI-10 item, compared to human response distribution. All models show systematic bias, underestimating high score [4-5] and overestimating low sores (0-2) for several items. Discrepancies were most pronounced for BFI-3, BFI-5, BFI-7, and BFI-10, with predictions errors reaching up to 200 response instances. For example, on BFI-10 ("has an active imagination"), the ground-truth frequency for high scores (4–5) was approximately 400, where GPT-4.1 Mini and Meta-LLaMA predicted fewer than 100 responses in these bins. In contrast, BFI-4 ("is relaxed, handles stress well") and BFI-9 ("gets nervous easily") showed relatively better alignment. DeepSeek tended to produce compressed distributions biased toward the lower-mid range (scores 1–3), often underrepresenting the upper end of the scale.

Figure 4 summarizes the exact match rates, off-by-one accuracy, and Cohen's kappa score across the BFI-10 items. Exact match rates ranged from 18% to 62%, with Meta-LLaMA achieving the highest exact match on BFI-1 ("is reserved", 62%), and GPT-4.1 Mini the lowest on BFI-10 ("has an active imagination", 18%). Off-by-one accuracy was substantially higher, with most values exceeding 70%. GPT-4.1 Mini reached 88% for BFI_5, and DeepSeek matched this on BFI_7. The lowest off_by_one accuracy was observed for Meta_LLaMA on BFI_3 (29%), followed by DeepSeek's on BFI_7 (39%). Despite moderate off-by-one accuracy, Cohen's kappa values were uniformly low, with most κ values below 0.10, such as BFI-5 and BFI-6, yielded near-zero or negative values, suggesting that the apparent agreement may be largely attributable to chance.

### LLM Prediction of Big Five Trait
Figure 5 summarizes model performance on the big five aggregated personality under zero-shot and chain-of-thought (CoT) prompting, using Pearson correlation, MAE, and RMSE. Across all traits and prompting conditions, correlation remained below r = 0.30. GPT-4.1 Mini demonstrated the best overall performance, with a peak of r= 0.25 for conscientiousness under zero-shot prompting. Conscientiousness was the only trait where any model exceeded r = 0.15. MetaLLaMA performed worst overall, with r=-0.10 for both extraversion and neuroticism (zero-shot prompting), and similarly poor performance under CoT.

Error metrics indicated MAE and RMSE generally exceeded 1.0, except for openness. Agreeableness exhibited the highest error, while openness showed the lowest. Meta-LLaMA recorded the highest overall errors (MAE = 1.69; RMSE = 1.95) on Agreeableness; GPT-4.1 Mini achieved the lowest errors for Openness (MAE = 0.54; RMSE = 0.73).

Figures 6 and 7 illustrate binned distribution (Low, Moderate, High) under Zero-shot and CoT prompting, respectively. All models tendered to overpredict Moderate and High scores, underestimating Low scores, particularly for Agreeableness and Conscientiousness, where predicted counts in the "High" bin exceeded ground-truth by nearly 300 (baseline ≈ 100). For Openness and Neuroticism, predictions were more closely aligned with ground-truth distribution. CoT prompting modestly improved distributional balance for Meta-LLaMA, which GPT-4.1 Mini and DeepSeek displayed negligible differences.

Figure 8 summarizes the exact match rates, off-by-one accuracy, and Cohen's kappa scores for Big Five prediction. Exact match rates ranged from 14% to 42%, with GPT-4.1 Mini (Zero-shot) achieving the highest rates for Openness (42%), for neuroticism (37%), and Extraversion (30%). The lowest match rates came from Meta-LLaMA (CoT) on Agreeable and Neuroticism (14%).

Off-by-one accuracy exceeded 70% in most conditions. Meta-LLaMA (CoT) reached 89% for Agreeableness and 81% for Neuroticism, while GPT-4.1 Mini (zero-shot) achieved 90% for openness. DeepSeek recorded the lowest rate (40%) for Agreeableness. Cohen's kappa remained low or negative, with the highest value observed for GPT-4.1 Mini (k = 0.058, zero-shot, Neuroticism), and the lowest for Meta-LLaMA (k = –0.044, Extraversion; –0.008, Openness; CoT), indicating minimal agreement beyond chance.

*Reliability Analysis*
Table 3 reports the Intraclass Correlation Coefficients (ICCs) for BFI-10 predictions across three independent runs of GPT_4o Mini. Most items showed high reliability (ICCs = 0.86-0.93). The highest value was for BFI-6 ("is outgoing, sociable") and BFI-8("does a thorough job"), both at 0.93. BFI_2 ("is generally trusting") and BFI_9 ("gets nervous easily) also showed excellent reliability (ICCs = 0.92 and 0.91, respectively). BFI_1("is reserved") and BFI_10 ("has an active imagination") followed closely to 0.90. The lowest reliabilities were observed for BFI-4 (0.88), BFI-7 (0.86), BFI-3 (0.75), and BFI-5 (0.61), the latter indicating relatively poor consistency.

Table 4 presents ICCs for the aggregated Big Five traits. All traits demonstrated strong to excellent reliability (ICCs > 0.80). Conscientiousness achieved perfect reliability (ICC = 1.00), followed by Extraversion (0.97), Neuroticism (0.95), Openness (0.88) and Agreeableness (0.81).

*Ablation Analysis: Impact of Input Context Length on Predictive Performance*
We conducted an ablation study using GPT_4o Mini to examine the impact of input context length (short: 100 tokens; medium: 1000 tokens; full context) on Big Five trait prediction. As show in Figure 9, medium-length input yields the highest Pearson correlations across traits, peaking for Agreeableness (r = 0.219) and Conscientiousness (r = 0.214). Short-input conditions produced substantially lower correlations, with most traits falling below *r* = 0.06 and a negative correlation for extraversion.

Figure 10 shows that longer input contexts increased RMSE. Full-context input resulted in the highest RMSE for Agreeableness (1.63), Conscientiousness (1.59), and Neuroticism (1.46). Short-input conditions produced the lowest RMSE for Extraversion (1.19) and Openness (0.789).

Figure 11 illustrates traits distribution across input lengths. Moderate bins aligned most closely with ground-truth. For Conscientiousness and Agreeableness, high scores were consistently overestimated and low scores underestimated. Openness was sharply overestimated (200+ responses vs. 50 ground truth). Similar trends were observed for Extraversion. Medium and full-context inputs showed better overall alignment.

Figure 12 presents exact match rates off-by-one accuracy by input length. Exact match rates improved modestly with longer inputs—especially for Openness (from 0.24 to 0.49) and Neuroticism (from 0.34 to 0.40). Off-by-one accuracy remained high (≥0.64) across all conditions.

Medium and full-context inputs consistently outperformed short input. The highest off-by-one accuracy was for Openness (0.94, medium; 0.96, full). Cohen's kappa values remained low (< 0.01) but slightly improved with longer inputs. The highest kappa was for Conscientiousness (κ = 0.088, full), while the lowest was for Openness (κ = –0.059, short).

## Discussion

This study evaluated the capacity of large language models (LLMs) to infer personality traits from semi-structured interviews using BFI-10 as a psychometric benchmark. While zero-shot prompting demonstrated high internal consistency, model prediction showed only modest alignment with ground-truth scores. GPT-4.1 Mini achieved the highest Pearson correlations and exact match rates, however, even its best-performing item, BFI_3 ("tends to be lazy"), reached only a modest Pearson correlation (r = 0.27). These finding underscores a persistent gap between linguistic fluency and psychometric validity in current LLMs. Although DeepSeek exhibited low numerical errors (MAE and RMSE), its weak and sometimes negative correlations suggest that numerical proximity alone does not imply meaningful psychological inference.

Beyond pointwise accuracy, our analysis revealed a systemic central tendency bias in model predictions. LLMs consistently underrepresented moderate trait ratings, disproportionately favoring extreme values. This effort was especially pronounced for abstract or socially nuanced traits like agreeableness and conscientiousness, while more concrete, behaviorally grounded items were predicted with relatively greater accuracy. These patterns indicate that LLMs may more effectively capture explicit behavioral signals than latent psychological constructs—a finding consistent with prior work on LLM-based trait inference from counseling or social media contexts [35,36].

Contrary to expectations, the incorporation of chain-of-thought (CoT) prompting showed limited benefit over zero-shot prompting. CoT prompting modestly improved categorical coherence for some models, particularly for Meta-LLaMA, yet failed to enhance correlation strength or reduce prediction error. This likely reflects the fact that personality traits are latent and globally distributed across language use, unlike fact-based tasks wher CoT excels. Personality inference may depend more on tone, affect, and latent semantic patterns than stepwise reasoning. The high off-by-one bin accuracy (>70%) indicates that models often approximate correct trait categories, even though exact predictions remain unreliable, as evidenced by uniformly low Cohen's kappa scores.

A key strength of LLM performance was its high test–retest reliability. Intraclass correlation coefficients (ICCs) exceeded 0.85 for most BFI-10 items and all Big Five traits, indicating strong output stability across repeated runs. Such consistency is essential for psychometric applications and provides a foundation for future model refinement.

Ablation analysis further revealed that increasing input context length modestly improved predictive correlations and categorical alignment but also lead to greater prediction variance. This trade-off highlights the importance of calibrating input length to optimize both performance and reliability.

Together, these findings suggest that current LLMs exhibit promising internal consistency and response stability but remain fundamentally constrained in their construct validity. By employing psychometrically validated self-report benchmarks rather than synthetic or heuristic labels, this study advances methodological rigor in the evaluation of LLM-based personality inference. Our finding resonate with recent critiques [37], underscoring the need for cautions, context-aware deployment of LLMs, especially in domains requiring impartial interpretation and psychological nuance.

Several limitations should be noted. Although the BFI-10 is psychometrically sound, its brevity may restrict expression of nuanced traits in text-based modeling. Moreover, none of the models were fine-tuned on domain-specific or psychologically annotated dialogue, which likely limited their inferential depth. Future work should explore fine-tuning on annotated conversational corpora, incorporating multimodal features (e.g., emotional tone, user profiler), and designing trait-specific prompting strategies to enhance both predictive accuracy and interpretability.

In conclusion, while current LLMs demonstrate fluency and reliability in generating personality predictions from interviews, their capacity to align with validated psychological constructs remains limited. These findings caution against overinterpretation of LLM-based trait inference and point toward necessary innovations in model training and prompt engineering. Our results contribute to the broader goal of developing responsible, evidence-based AI systems for psychological assessment and intervention.

## Materials and Methods

We describe the dataset, preprocessing procedures, prompting strategies, ablation design, and psychometric evaluation framework used to assess personality inference from naturalistic interview language.

*Dataset and Preprocessing*

We utilized a novel dataset comprised 555 U.S. adults recruited across multiple behavioral research studies focused on adjustment to adverse or transitional life events (e.g., occupational stress, health conditions). All participants provided written informed consent and engaged in the interview that was audio recorded for later transcription. Interviewers followed a script that included standardized prompts. Demographic information, including age and sex, was collected across studies and pooled for analysis. Participants had a mean age of 39.4 years (SD = 16.0; range: 18–89). The sample was approximately balanced by sex, with 278 participants identifying as female and 277 as male; one participant did not report their sex. A representative sample of participant responses across the five core questions is provided in Table 1.

Interview responses were preprocessed using a standard natural language processing pipeline. Text was lowercased, punctuation removed, and repeated words or filler phrases truncated. Following preprocessing, 518 participants with complete and valid Big Five Inventory–10 (BFI-10) scores were retained. The BFI-10 was administered during the same session as the interview, providing ground-truth scores on a 1 to 5 Likert scale for the five personality traits: conscientiousness, agreeableness, neuroticism, openness, and extraversion.

*Large Language Models and Prompting Strategies*

To assess LLM-based personality inference, we benchmarked three recent instruction-tuned models:

GPT-4.1Mini (OpenAI, 2025): A lightweight, high-performance proprietary model accessed via the OpenAI API [38].

Meta-LLaMA-3.3-70B-Instruct-Turbo: An instruction-tuned version of Meta's LLaMA 3.3 with 70B parameters, designed for alignment with human instruction following [39].

DeepSeek-R1-Distill-70B: A distilled variant emphasizing low-resource deployment while maintaining core language modeling abilities [40].

Two inference tasks were evaluated. For BFI-10 prediction, we employed a zero-shot prompting strategy: models received the participant's narrative and were asked to generate ten item-level BFI scores (1–5 scale) without being provided with examples or reasoning (see Tabel 5). This tested the LLMs' capacity to extract psychometric signals from open-ended language.

For Big Five inference, we compared zero-shot and Chain-of-Thought (CoT) prompting (see Tabe 6 and 7). In the zero-shot condition, models were prompted to directly generate a score (1-5) for each of the five broad dimension. In the CoT condition, prompts explicitly guided models through a multi-step reasoning process: first identifying relevant linguistic cues, then reflecting on their implications, and finally producing a synthesized trait score. This experimental design enabled us to assess whether structured reasoning enhances inference quality relative to direct prediction.

*Evaluation Metrics*

Model predictions were evaluated using a comprehensive psychometric framework that captures accuracy, consistency, and interpretability across continuous and categorical scoring formats.

**Criterion Validity:** Computing Pearson correlation coefficients between predicted and self-reported personality scores. This metric quantifies the linear agreement between model inferences and validated human ground truth and was applied separately to BFI-10 predictions and Big Five trait estimates under both prompting conditions.

**Prediction Error:** Mean Absolute Error (MAE) and Root Mean Squared Error (RMSE) captured average and variance-weighted deviations, respectively. MAE reflects the average magnitude of error regardless of direction, while RMSE provides a variance-weighted indicator that penalizes larger deviations.

**Categorical Alignment:** For self-reported BFI-10 data, we applied standard binning conventions from prior literature: Low (1–2), Moderate (3), and High (4–5) [41]. Since LLM-predicted scores are continuous, we discretized them using a coarser scheme: Low (1.0–2.4), Moderate (2.5–3.4), and High (3.5–5.0), to enable bin-wise comparison. From this, we computed (1) exact match rate (i.e., proportion of predictions that fall into the same bin as the ground truth), (2) off-by-one accuracy (i.e., within one adjacent bin), and (3) Cohen's kappa (κ), a chance-adjusted agreement statistic. Test–retest reliability: Using GPT_4o Mini to rerun model inference three times on a randomly selected subsample of 50 participants. Each run used a randomized ordering of input text. We then computed intraclass correlation coefficients (ICC [3, k]) to evaluate prediction stability. ICC values provide a robust index of the consistency of LLM outputs under minor perturbations, serving as a psychometric indicator of model reliability.

*Ablation Analysis: Input Length*

To evaluate how input context length influences model performance, we conducted an ablation analysis using GPT-4.1 Mini across three levels of transcript truncation. Specifically, we compared predictions generated from transcripts truncated to the first 100 words (response_short), the first 1,000 words (response_medium), and the full-length interview (response_full). The average length of the full transcripts was approximately 2,955 words (SD = 1,855). This analysis allowed us to assess how varying levels of contextual richness affect the accuracy, reliability, and categorical alignment of Big Five trait predictions. The statistic significant effects of input length on predicted trait values for all five Big Five dimensions (all $p < .001$): Conscientiousness ($\chi^2 = 86.71$), Agreeableness ($\chi^2 = 288.19$), Neuroticism ($\chi^2 = 177.63$), Openness ($\chi^2 = 466.25$), and Extraversion ($\chi^2 = 175.89$).

Model outputs were evaluated using continuous metrics (MSE, RMSE, Pearson correlation) and categorical metrics based on binning. Predicted scores were converted to ordinal categories (Low, Moderate, High) to examine trait-level alignment across input lengths.

**Figures and Tables**
**Table 1. Example Interview Responses**

| |
|---|
| **Participant ID: 001** <br> **Q1 (Daily Activities):** <br> "...Beginning of the day, uh I have two sons, one is [xxxx]. Spent the morning with them and, uh, kind of a slow morning. Towards the afternoon, I started to play outside for a half an hour, spent a little time at the gym..." <br> **Q2 (Recent difficult experience):** <br> "...Being a firefighter has been a challenging and amazing experience. I've been a firefighter since 2008, and a paramedic since 2004. Started right after high school when I was 18 years old. I've enjoyed it—it's been a lot of bad experiences too…" <br> **Q3 (Coping and emotional regulation):** <br> "...I think you're asking why I don't go places. The honest answer is, you talk to people you trust at work, but eventually... My wife and I have been married since I got on the job..." <br> **Q4 (Recent negative event):** <br> "...I was a newer lieutenant by 3 months when the first big incident happened. We went in as fast as possible. After we put most of the fire out, it was... stressful leading into it..." <br> **Q5 (Recent Positive event):** <br> "...Two of them always come to mind. The first one is the first baby we delivered—early morning call at 3 or 4 a.m.—it was a heroin case…" |

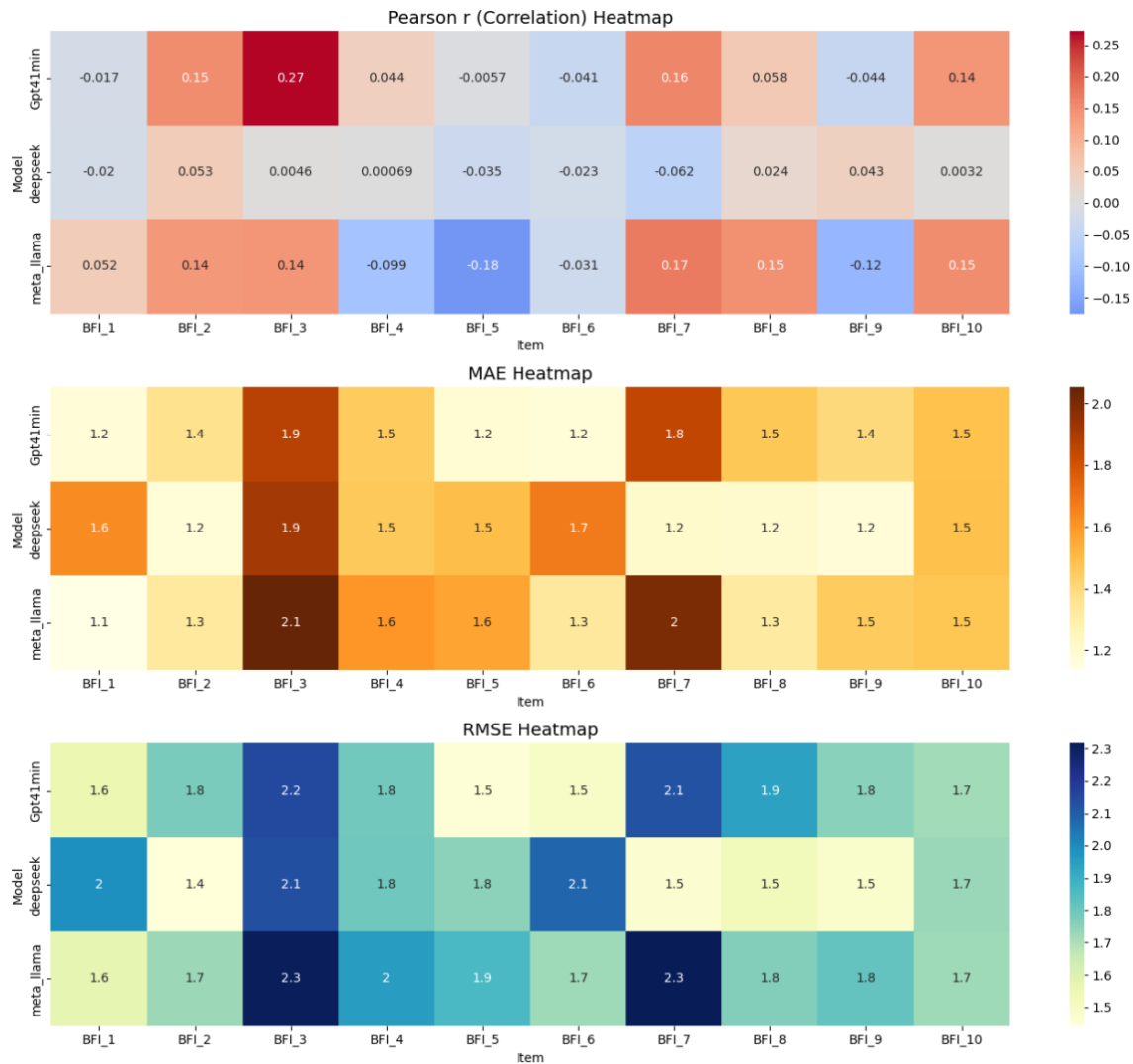

Figure 2. Comparative Performance of LLMs on BFI-10 Trait Prediction

**Table 2. Item-Level Prediction Performance and Interpretability for BFI-10 Traits**

| Item | Trait Description | Best Model | Best r | Worst Model | Worst r | Interpretation |
|---|---|---|---|---|---|---|
| **BFI_1** | Is reserved | Meta-LLaMA | 0.052 | DeepSeek | –0.020 | "Reserved" is abstract and context sensitive. It's often expressed through **absence** (short responses, passive tone), which is hard for LLMs to detect. |
| **BFI_2** | Is generally trusting | GPT-4.1 Mini | 0.15 | DeepSeek | 0.053 | Trust is a **latent socio-emotional trait**. Unless the user mentions openness, betrayal, or doubt, it's difficult to infer from neutral language. |
| **BFI_3** | Tends to be lazy | GPT-4.1 Mini | **0.27** | DeepSeek | 0.0046 | This is the **best predicted item**. It reflects **concrete, behavior-based self-evaluation**, often stated explicitly ("I procrastinate", "I avoid work"). |
| **BFI_4** | Is relaxed, handles stress well | GPT-4.1 Mini | 0.044 | Meta-LLaMA | –0.099 | Being "relaxed" is subtly expressed and context dependent. Without stress-related language or affective cues, models struggle to judge this trait. |
| **BFI_5** | Is full of energy | GPT-4.1 Mini | –0.0057 | Meta-LLaMA | –0.18 | This trait requires detection of **exuberant, enthusiastic expression**. Low |
| | | | | | | performance indicates users don't always convey this overtly. |
| BFI_6 | Is outgoing, sociable | DeepSeek | –0.023 | GPT-4.1 Mini | –0.041 | Despite being clear in meaning, models underperform — possibly because sociability requires **contextual clues** (e.g., group mentions, social events). |
| BFI_7 | Tends to find fault with others | Meta-LLaMA | 0.17 | DeepSeek | –0.062 | Negatively framed traits are harder to detect unless users express **judgmental or critical language**, which is rare unless explicitly prompted. |
| BFI_8 | Does a thorough job | Meta-LLaMA | 0.15 | DeepSeek | 0.024 | Being "thorough" reflects conscientious behavior that may be communicated through mentions of diligence, attention to detail, or responsibility. However, such traits are often implied rather than explicitly stated in open-ended responses. |
| BFI_9 | Gets nervous easily | DeepSeek | 0.043 | Meta-LLaMA | –0.12 | Emotional stability is often inferred from tone or emotional volatility. If users don't express emotional highs/lows, models can misinterpret. |
| BFI_10 | Has an active imagination | Meta-LLaMA | 0.15 | DeepSeek | 0.0032 | This trait is reflected in **vivid, metaphorical, or abstract language**. LLMs perform moderately here if the user demonstrates creativity in expression. |

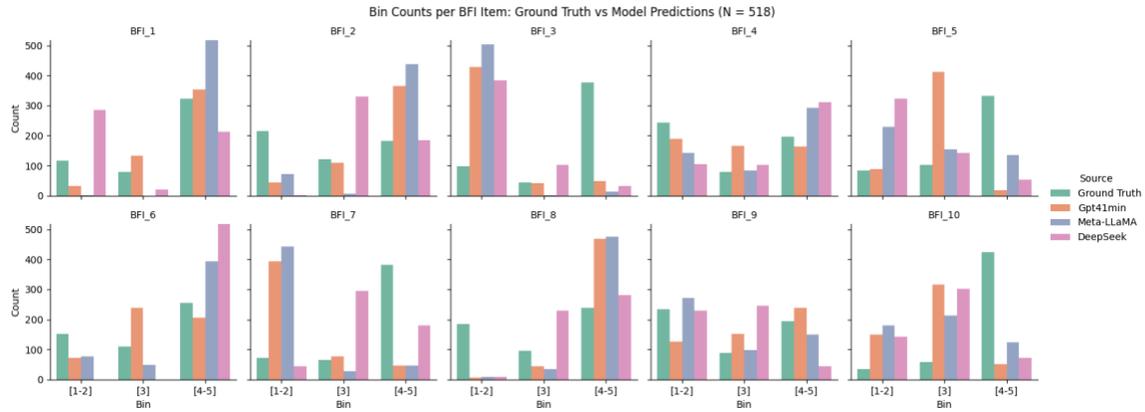

**Figure 3** Binned Distributions of BFI-10 Responses: Ground Truth vs. Model Predictions
Each panel represents one BFI-10 item, with bar heights indicating the number of responses in each Likert bin ([1–2], [3], [4–5]) across ground-truth data and predictions from three models (GPT-4.1 Mini, Meta-LLaMA, and DeepSeek). Compared to human responses, LLMs tend to underutilize the neutral category ([3]) and overpolarize predictions toward the extremes.

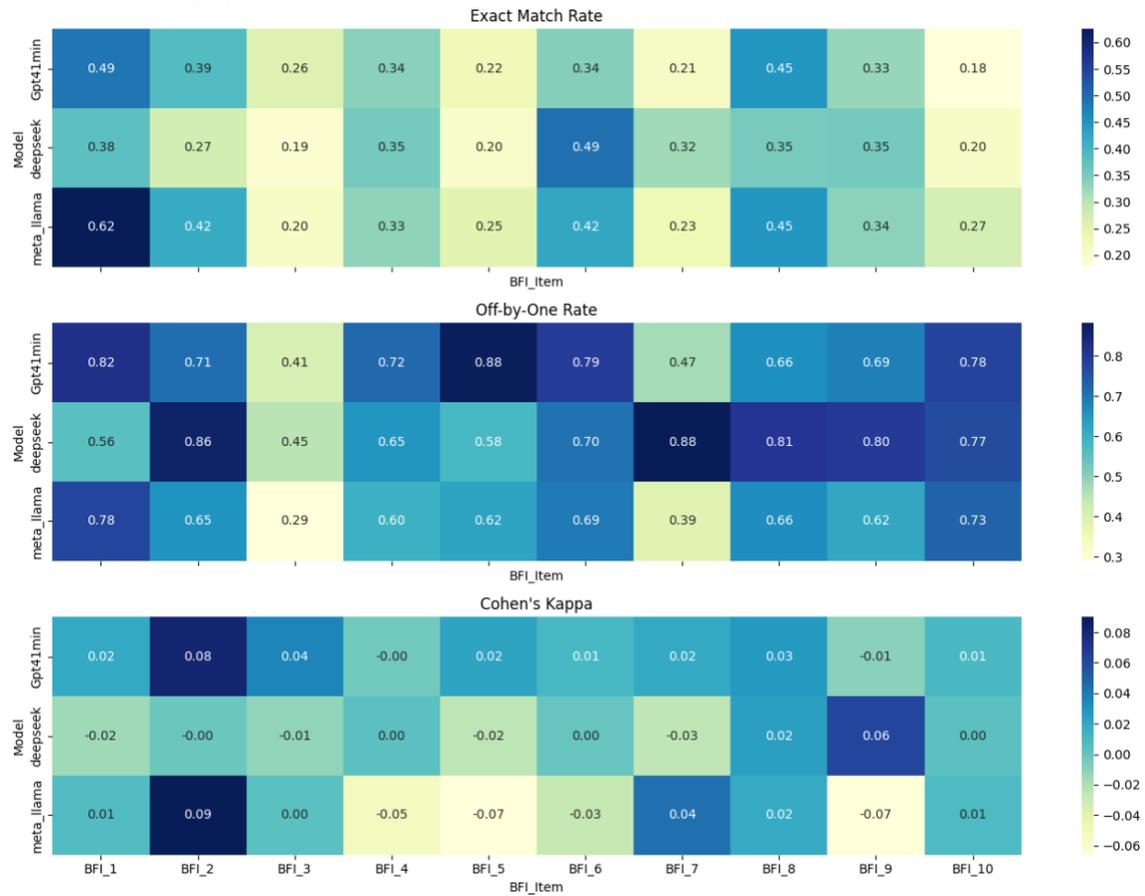

**Figure 4** Agreement Metrics Between Model Predictions and Ground-Truth BFI-10 Scores
This figure presents three agreement-based evaluation metrics across BFI-10 items for each model: Exact Match Rate (top), Off-by-One Rate (middle), and Cohen's Kappa (bottom). While Off-by-One accuracy is relatively high across all models, Cohen's Kappa reveals stronger item-level agreement for GPT-4.1 Mini, especially on BFI_3 and BFI_7. Meta-LLaMA shows moderate consistency, while DeepSeek demonstrates lower exact match and kappa agreement, suggesting less alignment with human-rated personality scores.

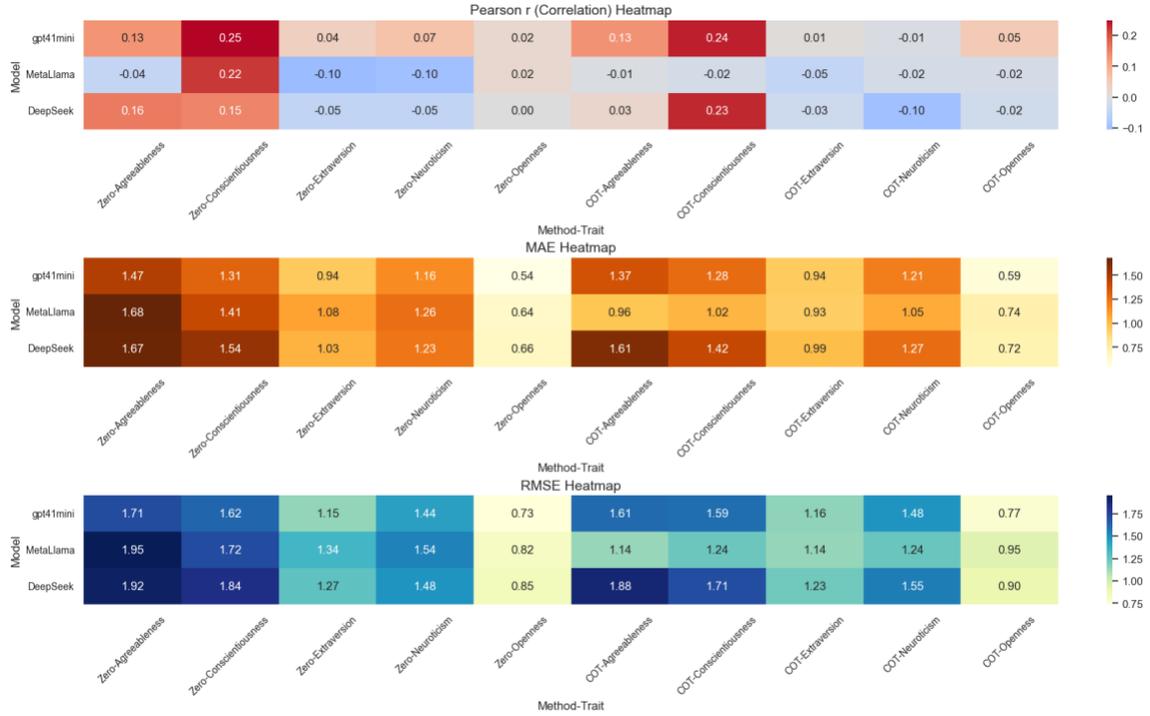

Figure 5 Comparative Performance of LLMs on Big Five Trait Prediction

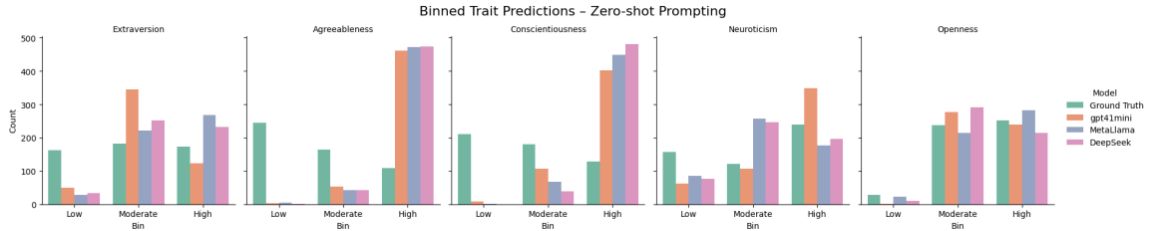

Figure 6 Binned Big Five Trait Predictions Using Zero-Shot Prompting Across Models

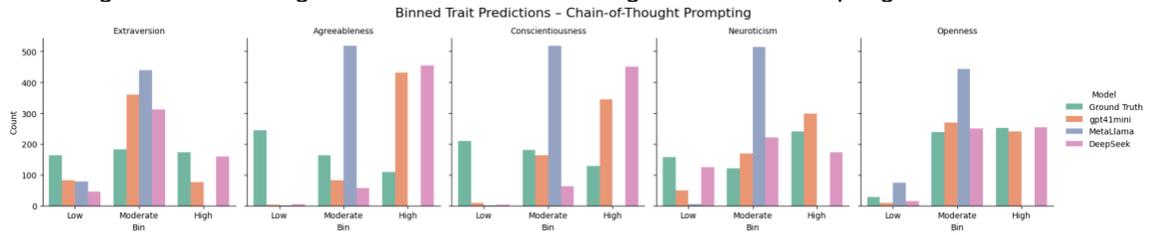

Figure 7 Binned Big Five Trait Predictions Using CoT Prompting Across Models

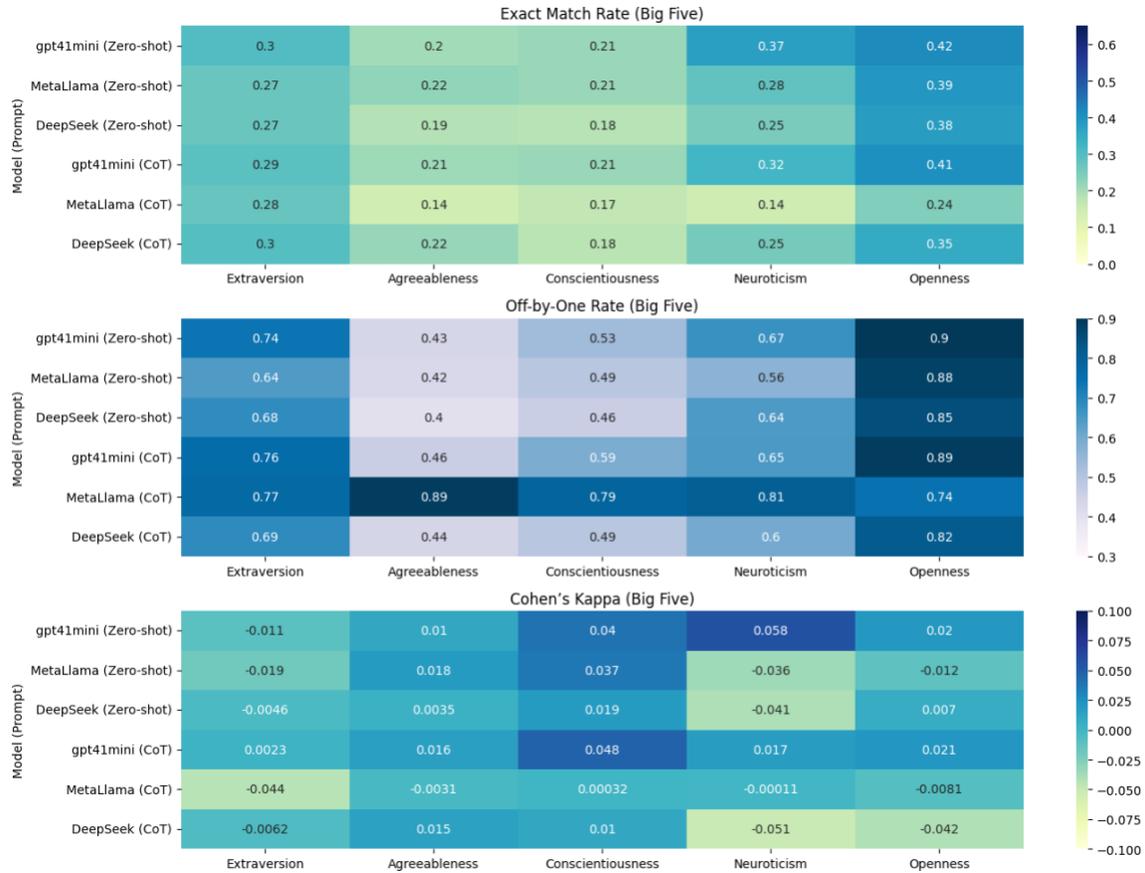

Figure 8 Categorical Agreement Metrics for Big Five Trait Predictions Across Models and Prompting Strategies

**Table 2** Best-Performing Models by Trait for Exact Match, Off-by-One Accuracy, and Cohen's Kappa

| Trait | Best Exact Match | Best Off-by-One | Best Kappa Score |
|---|---|---|---|
| Extraversion | GPT-4.1 Mini (Zero-shot) – 0.30 | MetaLlama (CoT) – 0.77 | GPT-4.1 Mini (CoT) – 0.002 |
| Agreeableness | MetaLlama (Zero-shot) – 0.22 | MetaLlama (CoT) – 0.89 | GPT-4.1 Mini (Zero-shot) – 0.01 |
| Conscientiousness | GPT-4.1 Mini / MetaLlama (Zero-Shot) – 0.21 | MetaLlama (CoT) – 0.79 | GPT-4.1 Mini (CoT) – 0.048 |
| Neuroticism | GPT-4.1 Mini (Zero-shot) – 0.37 | MetaLlama (CoT) – 0.81 | GPT-4.1 Mini (Zero-shot) – 0.058 |
| Openness | GPT-4.1 Mini (Zero-shot) – 0.42 | GPT-4.1 Mini (Zero-shot) – 0.90 | GPT-4.1 Mini (CoT) – 0.021 |

**Table 3.** Intraclass Correlation Coefficients (ICC) for BFI-10 Items Across Three Runs

| BFI Item | ICC |
|---|---|
| BFI_1 | 0.90 |
| BFI_2 | 0.92 |
| BFI_3 | 0.75 |
| BFI_4 | 0.88 |
| BFI_5 | 0.61 |
| BFI_6 | 0.93 |
| BFI_7 | 0.86 |
| BFI_8 | 0.93 |
| BFI_9 | 0.91 |
| BFI_10 | 0.90 |

**Table 4.** Intraclass Correlation Coefficients (ICC) for Big Five Traits Across Three Runs

| Personality Traits | ICC |
|---|---|
| Conscientiousness | 1.0 |
| Agreeableness | 0.81 |
| Neuroticism | 0.95 |
| Openness | 0.88 |
| Extraversion | 0.97 |

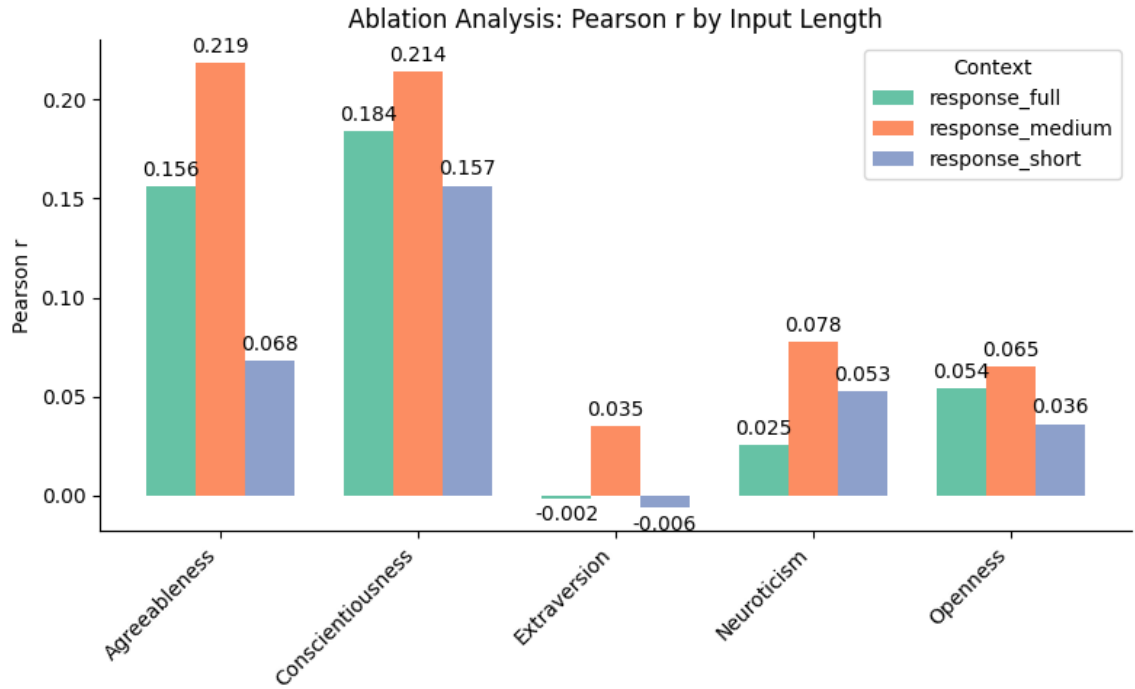

**Figure 9:** Effect of Context Length on Pearson Correlation Between Predicted and Ground-Truth

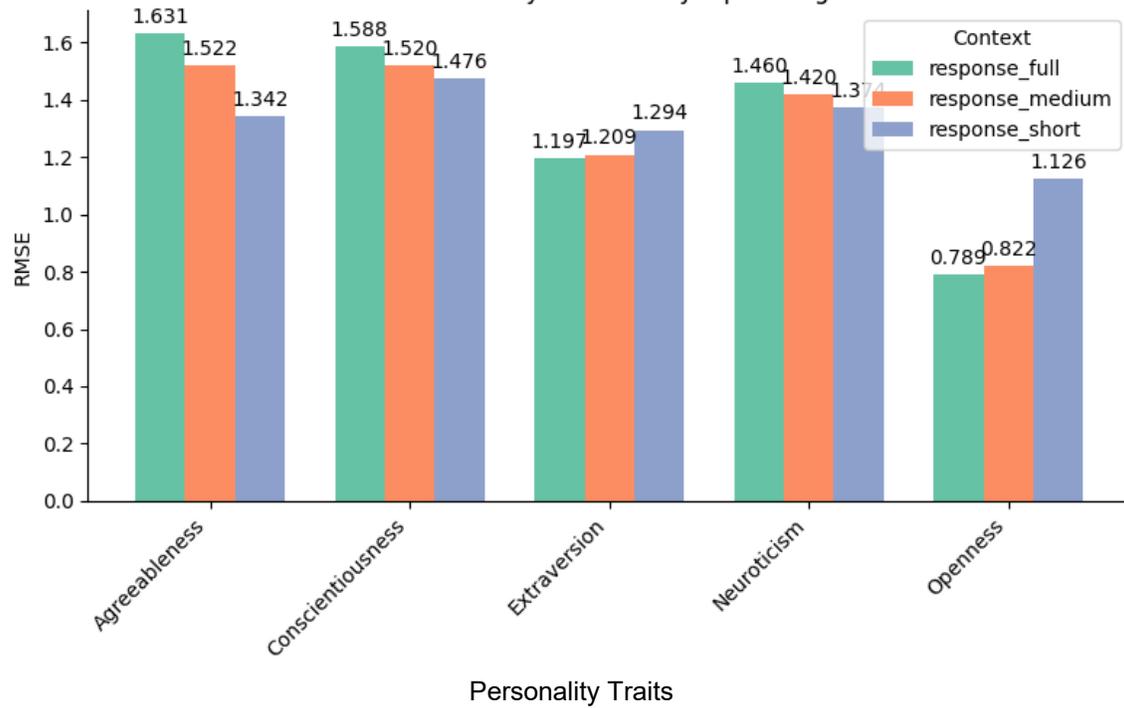

Personality Traits

**Figure 10:** Effect of Context Length on RMSE of Predicted Personality Trait Scores

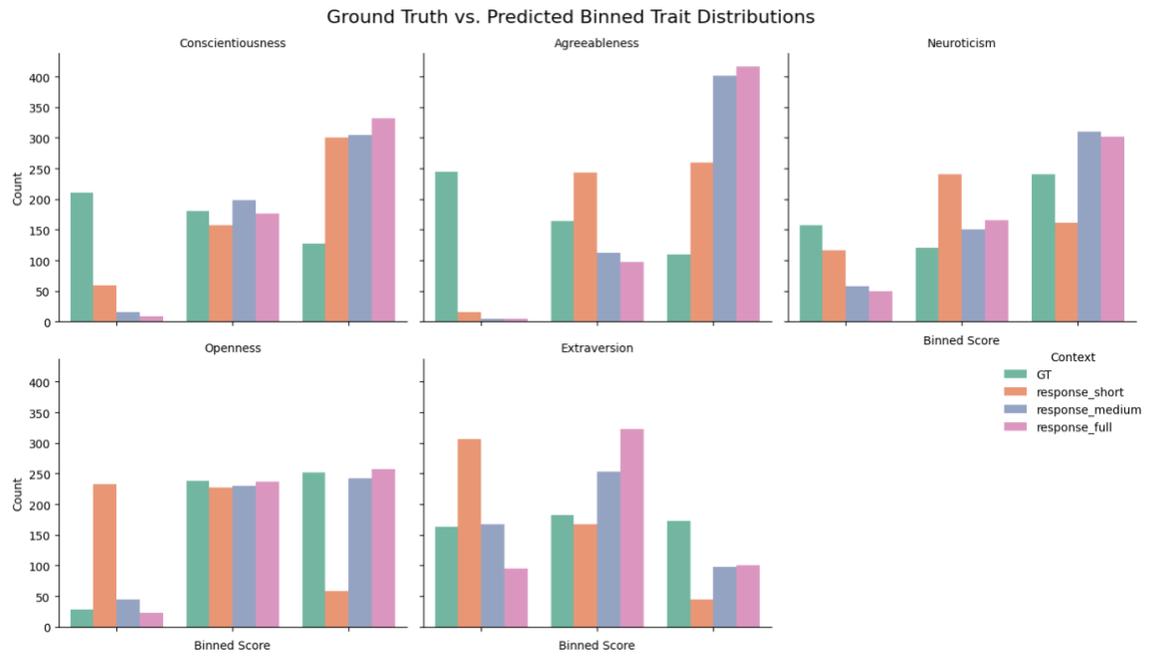

**Figure 11:** Ground Truth vs. Predicted Binned Trait Distributions by Input Context Length

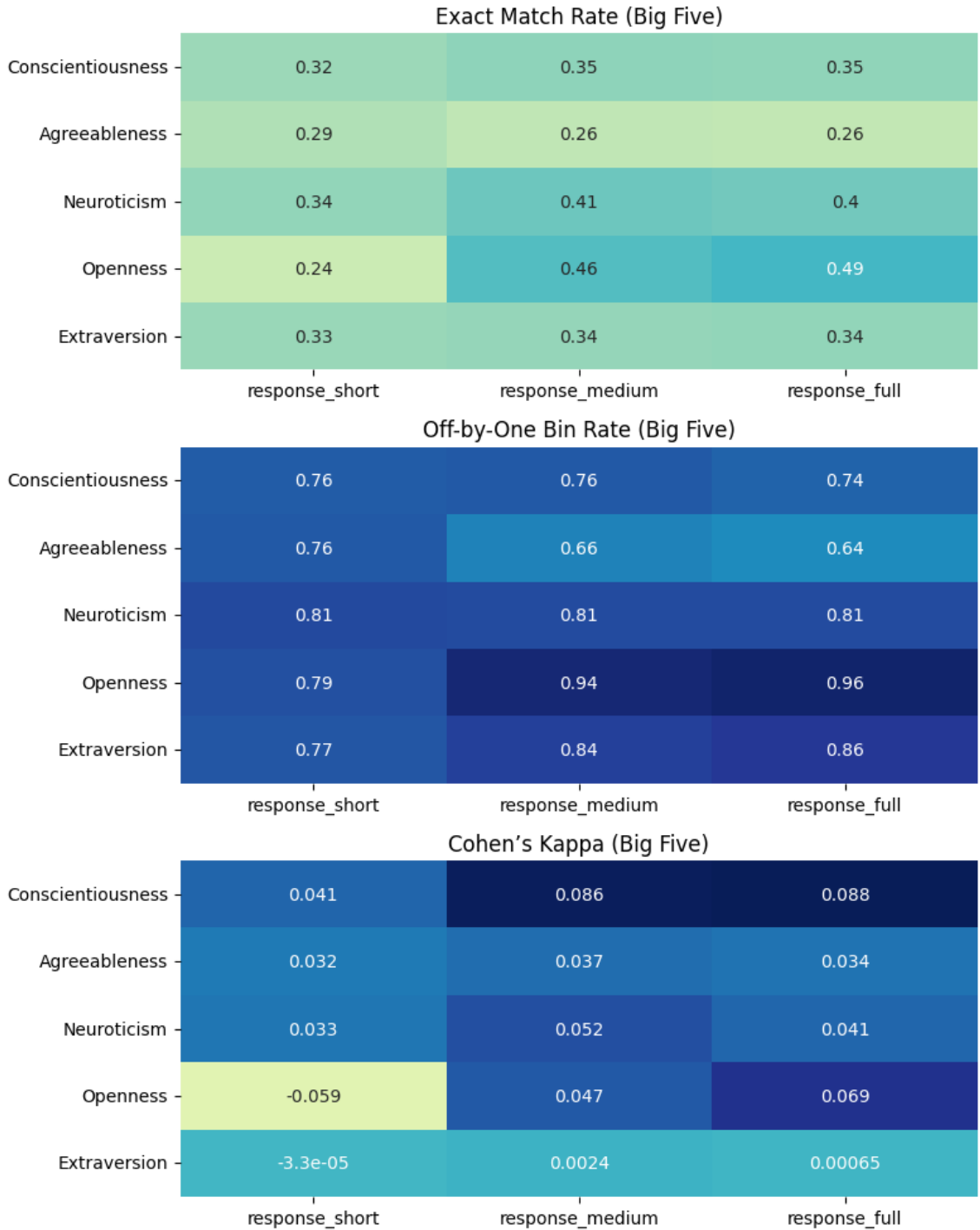

**Figure 12** Categorical Agreement Metrics for Big Five Trait Predictions Across Models with three response contexts

**Table 5.** Zero-shot prompt template for BFI-10 item-level personality inference

You are a clinical psychologist specializing in personality assessment. Based on the individual's response below, infer how they might score on each item from the Big Five Inventory (BFI-10).
Participant Response:
{response}
Your Task:
Carefully read the participant's narrative. Reflect on their attitudes, behaviors, and emotional tone. For each of the 10 BFI-10 items, consider how strongly that trait seems to describe the person.
Do not assume all traits are equally visible. Assign scores only based on what is *actually expressed*, using the following scale:
- 1 = Disagree strongly
- 2 = Disagree a little
- 3 = Neutral or unclear
- 4 = Agree a little
- 5 = Agree strongly
Focus on inference, not formula.
Output Format:
Return your personality scores in this format:
BFI_1_score: X  is reserved
BFI_2_score: X   is generally trusting
BFI_3_score: X   tends to be lazy
BFI_4_score: X   is relaxed, handles stress well
BFI_5_score: X   has few artistic interests
BFI_6_score: X   is outgoing, sociable
BFI_7_score: X   tends to find fault with others
BFI_8_score: X   does a thorough job
BFI_9_score: X   gets nervous easily
BFI_10_score: X   has an active imagination
Where X is a score from 1 to 5, derived from your qualitative interpretation of the participant's response.

**Table 6.** Zero-shot prompt template for Big Five item-level personality inference

You are a psychologically insightful agent. Given the following conversation response, first reflect on the speaker's language, behavior, and implicit attitudes that may signal stable personality traits based on the Big Five model: Conscientiousness, Agreeableness, Neuroticism, Openness, and Extraversion.
Step-by-step reasoning:
1. Summarize the emotional tone and key themes (e.g., positive, negative, confident, anxious).
2. Identify any linguistic cues or behaviors that indicate the traits (e.g., responsibility for conscientiousness, empathy for agreeableness).
3. Justify the trait scores based on these observations in 2-3 sentences.
Provide a score for each trait, from 1.0 to 5.0 (inclusive), allowing for half-point values such as 3.5. Use the exact format below:
Conscientiousness: X
Agreeableness: X
Neuroticism: X
Openness: X
Extraversion: X
Now analyze the following response:
\"\"\"{response}\"\"\"

**Table 7.** Chain-of-Thought prompt template for Big Five item-level personality inference

You are a psychologically insightful agent. Given the following conversation response, reflect on the speaker's language, behavior, and implicit attitudes that signal Big Five personality traits: Conscientiousness, Agreeableness, Neuroticism, Openness, and Extraversion.
Response:
\"\"\"{response}\"\"\"
Step-by-step reasoning:
1. Summarize the emotional tone and key themes (e.g., positive, negative, confident, anxious).
2. Identify any linguistic cues or behaviors that indicate the traits (e.g., responsibility for conscientiousness, empathy for agreeableness).
3. Justify the trait scores based on these observations in 2-3 sentences.
Provide a score (1.0-5.0) for each trait, allowing for half-step values like 3.5. Strictly use this format:
   Conscientiousness: X
   Agreeableness: X
   Neuroticism: X
   Openness: X
   Extraversion: X
   Where X is a number between 1.0 and 5.0 (inclusive), allowing for decimals such as 3.5.